\documentclass[conference]{IEEEtran}

\IEEEoverridecommandlockouts                              

\usepackage[dvipsnames, table]{xcolor}
\usepackage[hidelinks, colorlinks=true, citecolor=orange]{hyperref}
\usepackage{amsmath} 

\usepackage{amssymb}  
\usepackage{graphicx}
\usepackage{subcaption}
\usepackage{float}
\usepackage{placeins}
\usepackage[most]{tcolorbox}
\usepackage{booktabs}
\usepackage{adjustbox}   

\usepackage[dvipsnames]{xcolor}
\usepackage{colortbl}
\usepackage{mathrsfs}
\usepackage{tikz}
\usetikzlibrary{positioning}
\usepackage{xspace}
\usepackage{multicol,multirow}

\usepackage{amsthm}
\usepackage[outline]{contour}
\usepackage{soul}
\usepackage[]{mdframed}
\usepackage{threeparttable}
\usepackage{pifont}
\usepackage{wrapfig}
\usepackage{amsmath, amsfonts, amssymb}   
\usepackage{mathtools}
\usepackage{pifont}%

\usepackage{enumitem}
\usepackage{booktabs}
\usepackage{float}
\usepackage[linesnumbered,ruled,vlined]{algorithm2e}
\usepackage{bm}
\usepackage{subcaption}
\usepackage{makecell}
\usepackage[font=small]{caption}

\title{\LARGE \bf

\textit{Are LLMs The Way Forward?} A Case Study on LLM-Guided Reinforcement Learning for Decentralized Autonomous Driving
}

\author{
\IEEEauthorblockN{Timur Anvar, Jeffrey Chen, Yuyan Wang, Rohan Chandra}
\IEEEauthorblockA{{Dept. of Computer Science, University of Virginia}\\\texttt{\{tka8edd, fyy2ws, ntc8tt, rohanchandra\}@virginia.edu}}
}

\begin{document}
\maketitle
\thispagestyle{empty}
\pagestyle{empty}

\begin{abstract}


Autonomous vehicle navigation in complex environments such as dense and fast-moving highways and merging scenarios remains an active area of research. In the past decade, many planning and control approaches have used reinforcement learning (RL) with notable success. However, a key limitation of RL is its reliance on well-specified reward functions, which often fail to capture the full semantic and social complexity of diverse, out-of-distribution situations. As a result, a rapidly growing line of research explores using Large Language Models (LLMs) to replace or supplement RL for direct planning and control, on account of their ability to reason about rich semantic context. However, LLMs present significant drawbacks: they can be unstable in zero-shot safety-critical settings, produce inconsistent outputs, and often depend on expensive API calls with network latency. This motivates our investigation into whether small, locally deployed LLMs ($\leq$14B parameters) can meaningfully support autonomous highway driving through reward shaping rather than direct control. These models are attractive for practical deployment as they can run on a single GPU and avoid external API dependencies. We present a case study comparing RL-only, LLM-only, and hybrid approaches, where LLMs augment RL rewards by scoring state-action transitions during training, while standard RL policies execute at test time. Our findings reveal that RL-only agents achieve moderate success rates (73-89\%) with reasonable efficiency, LLM-only agents can reach higher success rates (up to 94\%) but with severely degraded speed performance, and hybrid approaches consistently fall between these extremes. Critically, despite explicit efficiency instructions, LLM-influenced approaches exhibit systematic conservative bias with substantial model-dependent variability, highlighting important limitations of current small LLMs for safety-critical control tasks.

\end{abstract}

 
\section{Introduction}
Highway driving is a critical benchmark for autonomous navigation. Unlike structured urban settings, highways combine high speeds, dense traffic, and frequent, often uncoordinated merges. Robust control in this setting must be both effective and socially aware by maintaining appropriate speeds, merging smoothly, respecting implicit norms, and avoiding aggressive maneuvers. These challenges are amplified in decentralized scenarios where vehicles must reason about others without direct communication~\cite{wu2023intent, chandra2022game, chandra2022gameplan, chandra2020cmetric, chandra2020graphrqi, chandra2021using, chandra2022towards, chandra2023meteor}.

Reinforcement learning (RL) has been extensively studied as a method to train agents to navigate complex environments~\cite{mavrogiannis2022b, haydari2022}. In driving applications, RL enables vehicles to learn behaviors such as lane changing, braking, and overtaking purely through interaction with the environment~\cite{chen2024deep, chandra2019traphic, chandra2020forecasting, chandra2019robusttp}. However, deploying RL in decentralized, partially observable highway settings remains a challenge. Agents face incomplete information~\cite{chandra2019densepeds, chandra2019robusttp}, must generalize across diverse traffic conditions~\cite{kothandaraman2021bomudanet, kothandaraman2021ss}, and rely on carefully engineered reward functions to encourage safety and cooperation~\cite{bouton2020improving}. Given the multifaceted demands of autonomous driving, which include safety, liveness, fairness, and social compliance, manually crafting reward functions that capture nuanced driving behaviors is difficult~\cite{zhou2024, suriyarachchi2022gameopt, suriyarachchi2024gameopt+}. Meanwhile, large language models (LLMs) have emerged as powerful tools capable of encoding rich, human-like reasoning~\cite{hagendorff2023human, song2025reward}. Trained on vast textual data, LLMs incorporate broad common sense knowledge and high-level behavioral expectations. However, LLMs alone can produce suboptimal or inconsistent decisions in safety-critical contexts due to hallucinations and lack of grounding, necessitating verification and control~\cite{huang2023hallucination}.

Given these complementary strengths and limitations of RL and LLMs, a natural question emerges: can we effectively combine the learning capabilities of RL with the semantic reasoning of LLMs for autonomous driving? Although large-scale LLMs have shown promise in various applications, their deployment presents significant practical barriers. They require substantial computational resources, incur per-token API costs, and introduce network dependencies that can compromise real-time performance in safety-critical applications. Such challenges naturally lead to the investigation of smaller, locally deployable alternatives. Small local LLMs ($\leq$14B parameters) offer compelling advantages: they operate on a single GPU, eliminate data sharing concerns and per-token costs, and avoid API rate limits and network latency issues. From a research accessibility perspective, these models are particularly important as they represent the computational budgets of typical academic laboratories, unlike the massive infrastructure required for frontier models from companies such as OpenAI.

However, this practical appeal raises the fundamental question: what can we realistically expect from these resource-constrained models in complex, safety-critical tasks such as autonomous driving? Do smaller LLMs retain sufficient semantic reasoning capabilities to meaningfully improve RL performance, or do their computational limitations fundamentally compromise their effectiveness?

\subsection{Main Contributions}
To address these questions, we conduct a systematic case study examining the capabilities and limitations of small local LLMs when integrated with RL for highway navigation. Rather than proposing a novel framework, our goal is to understand what works, what fails, and why. We examine a hybrid approach where small local LLMs are used during RL training  for reward shaping, semantically scoring state-action transitions, while the deployed policy remains a standard RL agent with no runtime LLM dependencies. This design preserves real-time efficiency at deployment while leveraging semantic guidance during learning.

The contributions of our case study are as follows:
\begin{enumerate}
    \item \textbf{Method Evaluation:} We compare three setups under a fixed prompt: an RL-only baseline, an LLM-only action policy, and a Hybrid (RL+LLM shaping) agent.
    \item \textbf{Practical Local Models:} We use Qwen3-14B and Gemma3-12B, both open-weight and practical on consumer hardware. Qwen3 reports that dense models (including 14B) improve reasoning, math, and coding, matching or even outperforming larger Qwen2.5 models, with explicit support for local deployment \cite{qwen3_blog}. The Gemma-3-12B model card reports competitive results on reasoning and STEM benchmarks and highlights efficient inference and deployment on laptops, desktops, and other resource-limited environments \cite{gemma_2025}. Together, they represent a broader class of efficient, high-performance midsize LLMs available for on-device use.


    \item \textbf{Varied Reward Shaping:} We examine three shaping schemes: dense (additive), averaged (equal-weight fusion with the environment reward), and centered (mean-zero adjustment). These represent a spectrum of ways to integrate semantic feedback: direct additive influence, balanced fusion, and bias-corrected guidance. These three schemes cover the main design axes of scale, weight, and centering, providing simple and interpretable comparisons appropriate for a case study.
    \item \textbf{Scenario Diversity:} We evaluate performance in multiple decentralized highway scenarios (\texttt{highway}, \texttt{highway-fast}, and \texttt{merge}) to test robustness under different traffic dynamics. Performance is measured using three interpretable metrics: collision-free success rate, mean number of lane changes, and mean speed score.
\end{enumerate}

\subsection{Preview of Findings}

Across 100-episode evaluations, RL-only agents attain success (collision-free) rates of $73\!-\!89\%$ with higher speeds, while LLM-only agents reach higher success rates of up to $94\%$, but with severely degraded speed performance. Hybrid approaches improve safety over RL in several settings but reduce efficiency, with Gemma3-12B generally producing safer yet slower behavior than Qwen3-14B.
Overall, our results indicate that small local LLMs can provide useful shaping signals but introduce a systematic conservative bias, a theme we analyze in Sections~\ref{Hybrid Design} and \ref{Discussion & Conclusion}.


\section{Related Work}
\label{Related Work}

\subsection{Deep Reinforcement Learning for Autonomous Driving}
Deep Reinforcement Learning (DRL) has become a promising paradigm for decision-making in autonomous driving~\cite{haydari2022}, encompassing tasks from lanekeeping to multi-agent coordination~\cite{liu2025} and intersection navigation~\cite{LI2022103452}. DRL approaches have shown strong performance in simulation environments, learning effective driving behaviors through environment interaction. However, some challenges hinder their scalability and reliability in real-world applications~\cite{xu2025}. First, DRL requires a lot of data, making deployment challenging and reduces scalability, especially for complex multi-agent driving scenarios~\cite{kiran2022}. Additionally, DRL models can be difficult to interpret~\cite{zhu2022}, undermining their ability to handle rare, out of distribution events.

Reinforcement Learning from Human Feedback (RLHF) has been explored as a way to mitigate these issues by incorporating expert knowledge~\cite{sun2023}. While RLHF can improve policy robustness by guiding learning with human intuition, it requires substantial human annotation~\cite{yu2024}. Additionally, human-generated feedback may not cover the full range of driving conditions, reducing its ability to generalize. These limitations underscore the need for alternate methods, such as LLMs, that offer the benefits of expert guidance without such high resource overhead~\cite{xu2025}.

\subsection{LLMs in Autonomous Driving and Decision-Making}
LLMs have also been studied in the context of autonomous driving, where their vast linguistic and reasoning capabilities can enhance decision-making~\cite{zhou2024}. Prompt engineering-based approaches allow LLMs to interpret driving scenarios and assist with decision-making in a closed-loop setting~\cite{fu2023}, sometimes with added reflection modules to support iterative reasoning~\cite{wen2024}. Conversely, fine-tuning pretrained LLMs with driving-specific data has also been explored~\cite{chen2023}. While this approach can improve decision quality, it comes at the cost of significant training data and computational overhead. Furthermore, LLMs tend to produce outputs with inherent randomness, leading to unpredictable behavior~\cite{xu2025}, which can be a problematic for autonomous vehicles that require safe, repeteable, deterministic actions. These limitations have motivated hybrid frameworks that leverage the reasoning power of LLMs while mitigating their drawbacks during real-time execution~\cite{zhou2024,xu2025}.

\subsection{Hybrid DRL-LLM Approaches and Reward Generation}

Recent research has begun exploring the combination of LLMs with DRL~\cite{xu2025}, as LLMs have increasingly been considered for integration with RL agents to support high-level reasoning tasks~\cite{zhou2024}. More research has sought to utilize RL to fine-tune LLMs~\cite{bai2022}, while the reverse of utilizing LLMs to assist RL agents remains less explored~\cite{xu2025}. Early work has explored utilizing LLMs to improve the exploration efficiency and learning effectiveness of RL agents~\cite{liu2024}, while other work has explored using LLMs to augment the skill sets for decision-making of RL agents~\cite{yuan2023}. Within this broader context, a growing area of research explores using LLMs as proxies for reward signals~\cite{zhou2024}, providing an alternative to traditional hand-crafted or human-curated reward functions. In particular, prior work has shown that LLMs can assign reward signals in social interaction tasks such as negotiation, relying solely on LLM-generated feedback~\cite{kwon2023}. Other approaches have explored generating and optimizing rewards directly through Python code generation using LLMs~\cite{ma2024}.

\section{Experimental Setup and Evaluation Framework}
\label{Experimental Setup}

To facilitate comparison across different approaches while avoiding repetition, we first establish the common experimental framework used throughout this study. All three approaches (RL-only, LLM-only, and hybrid) share the same environment, observation space, action space, and evaluation metrics.

\subsection{Experimental Setup}
The experiments were carried out on a machine running Ubuntu 24.04, equipped with an AMD Ryzen 7 7800X3D CPU and an AMD Radeon RX 7900 XTX GPU. RL training and environment simulations were implemented in Python using Stable Baselines3 and \texttt{highway-env}~\cite{highway-env}, while LLM evaluations were performed locally with Qwen3-14B and Gemma3-12B via the Ollama framework.

The ego agent observes a compact 4-dimensional time to collision (TTC)-based vector containing ego speed and TTC values for the left, current, and right lanes. The agent selects from five discrete meta-actions: \texttt{LANE\_LEFT}, \texttt{IDLE}, \texttt{LANE\_RIGHT}, \texttt{FASTER}, and \texttt{SLOWER}. We implement a custom wrapper that converts raw environment outputs into this representation and integrates LLM reward shaping during training. We train our agents on the \texttt{highway-fast} environment, which offers reduced computational overhead while maintaining similar task complexity. For evaluation, each model is tested on 100 episodes per environment, with the random seed set to the episode number to ensure reproducibility. This controlled setup enables direct comparison of the three approaches while evaluating the role of small, local LLMs in autonomous driving tasks.

\subsection{Evaluation Metrics}
We report three aggregate metrics (Success Rate, Lane-Change Score, and Speed Score) over $N$ evaluation episodes, with $N=100$ by default. Success Rate (SR) is defined as 
$\mathrm{SR} = \frac{1}{N}\sum_{e=1}^{N} y_e \times 100$, 
where $y_e \in \{0,1\}$ indicates whether episode $e$ finishes without a collision. Lane-Change Score (LC) is given by 
$\mathrm{LC} = \frac{1}{N}\sum_{e=1}^{N} c_e$, 
where $c_e$ denotes the number of lane changes in episode $e$. We calculate Speed Score in two steps. First, we calculate the mean speed across all episodes, 
$\bar v = \frac{1}{N}\sum_{e=1}^{N} \bar v_e$, 
where $\bar v_e$ is the mean ego speed in episode $e$. Next, we normalize this value using the lower and upper bounds $v_{\min}=20$ m/s and $v_{\max}=30$ m/s, resulting in 
$\mathrm{Speed} = \mathrm{clip}\!\left(\frac{\bar v - v_{\min}}{v_{\max} - v_{\min}},\,0,\,1\right)$. Generally, Speed Scores closer to $1$ means the agent maintained a faster than average speed most of the times.

\section{Case Study Between RL-only and LLM-only Autonomous Driving}
\label{Problem Formulation}

\subsection{RL-only Problem Formulation}
We consider the problem of autonomous highway navigation for a single ego agent operating within a multi-agent environment populated by other independent vehicles. These surrounding vehicles behave according to fixed or scripted policies, with no direct communication or coordination between them. The ego agent must make safe, efficient, and socially compliant driving decisions based solely on its local observations, without access to the full global state.

This scenario is naturally modeled as a partially observable Markov decision process (POMDP), defined by the tuple, $\left \langle X, U, T, \Omega, O, R, \pi \right \rangle$, where the state space $X$ represents the full underlying state of the environment, and the action space $U$ is the discrete set of driving actions available to the agent. The transition function $T: X \times U \rightarrow X$ defines the mapping from current state and action to the next state, while the observation space $\Omega$ contains the set of observations accessible to the agent, providing partial and local views of the environment. The observation function $O: X \times U \rightarrow \Omega$ maps state and action to an observation, and the reward function $R: X \times U \rightarrow \mathbb{R}$ provides scalar feedback based on the desirability of state-action pairs. Finally, the policy $\pi: \Omega \rightarrow U$ maps observations to actions. At each timestep, the agent receives an observation from \(\Omega\), selects an action from \(U\) according to \(\pi\), and transitions to the next state according to \(T\). The agent’s objective is to maximize the expected discounted sum of rewards over time. We adopt this standard POMDP formulation to provide a structured foundation for a case study of how local LLMs can influence agent learning when incorporated as shaping signals.


\begin{figure}[tb]
\centering
\begin{tcolorbox}[colback=blue!5, colframe=blue!40!black, title=DQN Hyperparameters]
\begin{itemize}
    \item Policy: MLP with two hidden layers, each of 256 units
    \item Learning rate: \(1 \times 10^{-4}\)
    \item Replay buffer size: 50,000
    \item Training starts after: 1,000 environment steps
    \item Batch size: 64
    \item Discount factor \(\gamma\): 0.98
    \item Training frequency: every 4 environment steps
    \item Gradient steps per training call: 4
    \item Target network update interval: every 1,000 steps
\end{itemize}
\end{tcolorbox}
\caption{DQN hyperparameters used in our RL-only experiments.}
\label{fig:rl-only_hyperparams}
\vspace{-10pt}
\end{figure}

\subsection{RL-only Results}
\begin{figure}[t]
    \centering
    \includegraphics[width=\columnwidth]{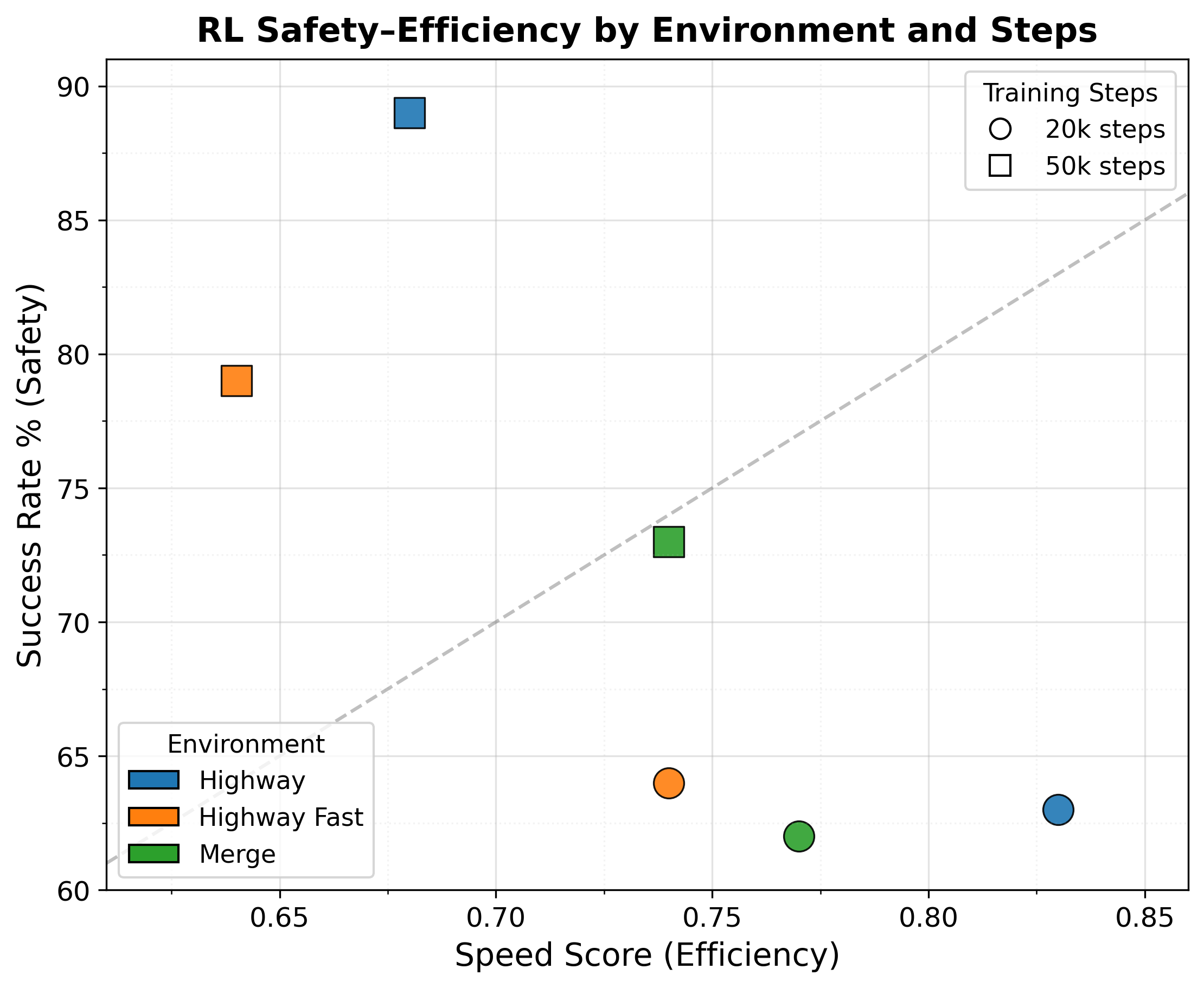}
    \caption{RL-only safety–efficiency spectrum. Longer training increases safety while reducing speed.}
    \label{fig:rl-tradeoff}
    \vspace{-10pt}
\end{figure}

We trained the baseline DQN agent using the experimental setup described in Figure~\ref{fig:rl-only_hyperparams} with a fixed random seed of 42. The agent was trained on both 20,000 and 50,000 timesteps to examine performance evolution. Table~\ref{tab:rl_combined_metrics} presents the evaluation results across three environments (\texttt{highway}, \texttt{highway-fast}, and \texttt{merge}) for both training durations. With additional training, the success rates increase substantially in all cases: for example, on \texttt{highway} the success rate increases from $63\%$ to $89\%$. This safety improvement is accompanied by a notable reduction in lane change frequency, indicating that longer training produces more stable and cautious driving policies. At the same time, the speed score decreases across environments (e.g., from $0.83$ to $0.68$ on \texttt{highway}), suggesting that gains in safety come partly at the expense of efficiency (see Fig.~\ref{fig:rl-tradeoff}). 


Together, these results provide a reference point for evaluating LLM-shaped agents. The overall trade-offs highlight key shortcomings of the RL-only approach. The slower and more conservative behavior that emerges with longer training shows RL's difficulty in balancing multiple competing objectives without carefully tuned reward functions. This motivates the exploration of other schemes, such as LLM-based reward shaping, to augment RL agents.

\begin{table}
\begin{center}
\begin{tabular}{ccccc}
    \toprule
    Environment & Steps & SR (\%) & Avg LC & Speed \\
    \midrule
    Highway & 20k & 63 & 9.01 & 0.83 \\
    Highway & 50k & 89 & 4.47 & 0.68 \\
    Highway-fast & 20k & 64 & 7.68 & 0.74 \\
    Highway-fast & 50k & 79 & 3.61 & 0.64 \\
    Merge & 20k & 62 & 2.76 & 0.77 \\
    Merge & 50k & 73 & 1.07 & 0.74 \\
    \bottomrule
\end{tabular}
\caption{Evaluation of RL-only agents across environments and training durations. 
Success rates rise substantially with more training steps, while lane changes decrease and speed scores decline, 
illustrating the safety–efficiency trade-off that motivates comparison with LLM-influenced approaches.}
\label{tab:rl_combined_metrics}
\end{center}
\end{table}

\subsection{LLM-Only Problem Formulation}
In contrast to the RL setup, the LLM-only agent does not optimize a reward function within the environment.  
Instead, its policy is directly induced by a pretrained language model.  
Formally, we define the tuple $\left\langle
(\Omega, U, f_{\text{LLM}}, \pi_{\text{LLM}}), \right \rangle$ where the observation space $\Omega$ matches the same space used in the RL setup, providing partial and local information about the driving environment. The action space $U$ consists of the same discrete meta-actions available to the RL agent. The LLM mapping $f_{\text{LLM}}$ is a fixed function that takes an observation $o_t$, formats it as a textual prompt, and produces an action preference. The policy $\pi_{\text{LLM}}$ selects the action 
\[
u_t = \pi_{\text{LLM}}(o_t) = f_{\text{LLM}}(\text{prompt}(o_t)).
\]

At each timestep, this policy translates the current observation into a textual prompt, queries the LLM to select one of the five discrete meta-actions (\texttt{LANE\_LEFT}, \texttt{IDLE}, \texttt{LANE\_RIGHT}, \texttt{FASTER}, \texttt{SLOWER}), and relies on the highway simulator's underlying proportional and proportional-derivative controllers to execute the corresponding low-level steering and acceleration commands. 


\begin{figure}[tb]
\centering
\begin{tcolorbox}[colback=blue!5, colframe=blue!40!black, title=LLM Action Prediction Prompt, enhanced jigsaw, breakable=false]

You are controlling an autonomous vehicle. Prioritize \textbf{safety first}, then consider efficiency and human-like behavior.

\vspace{0.5em}
\textbf{Guidelines:}
\begin{enumerate}[leftmargin=2em]
    \item Avoid any action that risks a collision.
    \item Treat any Time-To-Collision (TTC) below 2 seconds as dangerous.
    \item Speed up only if the current lane TTC (center lane) is clearly safe (above 3 seconds).
    \item If the current lane TTC is low (below 2 seconds), and either left or right lane TTC is high (above 3 seconds), prefer changing to the safest lane.
    \item Maintain a target speed around 30 m/s if safe, but slowing down is acceptable when unsure.
    \item Change lanes only if it clearly improves safety or avoids slower traffic.
\end{enumerate}

\vspace{0.5em}
\textbf{Available actions:}
\begin{center}
\begin{tabular}{ll}
0 = Turn Left & 1 = Idle \\
2 = Turn Right & 3 = Go Faster \\
4 = Slow Down & \\
\end{tabular}
\end{center}

\vspace{0.5em}
\textbf{Current observations:}
\begin{itemize}[leftmargin=2em]
    \item Speed: \{ego\_speed:.1f\} m/s
    \item Left lane TTC: \{left\_ttc:.1f\} s
    \item Center lane TTC: \{current\_ttc:.1f\} s
    \item Right lane TTC: \{right\_ttc:.1f\} s
\end{itemize}

What is the safest and most reasonable driving action to take now? Respond with only the action number (0--4).

\end{tcolorbox}
\caption{Text prompt used in our LLM-only experiments.}
\label{fig:llm-only_prompt}
\end{figure}

\subsection{Results for LLM-only Agent}
\begin{table}[t]
\begin{center}
\begin{tabular}{ccccc}
    \toprule
    Environment & LLM & SR (\%) & Avg LC & Speed \\
    \midrule
    Highway & Qwen3-14B & 87 & 0.06 & 0.23 \\
    Highway & Gemma3-12B & 91 & 0 & 0.07 \\
    Highway-fast & Qwen3-14B & 72 & 0.03 & 0.26 \\
    Highway-fast & Gemma3-12B  & 94 & 0 & 0.05 \\
    Merge & Qwen3-14Bs & 31 & 0.01 & 0.56 \\
    Merge & Gemma3-12B  & 75 & 0 & 0.14 \\
    \bottomrule
\end{tabular}
\caption{Evaluation of LLM-only agents across environments. 
Models achieve competitive success rates, often exceeding RL baselines, 
but with near-zero lane changes and severely reduced speed scores, 
highlighting their conservative driving bias. Notably, the fastest-performing 
LLM configuration corresponds to the lowest success rate.}
\vspace{-10pt}
\label{tab:llm_combined_metrics}
\end{center}
\end{table}

\begin{table}[t]
\centering
\begin{tabular}{lccc}
\toprule
 & \textbf{Faster} & \textbf{Idle/Faster} & \textbf{Right Lane} \\
\midrule
Ego Speed (m/s) & 25.0 & 28.0 & 22.0 \\
Left Lane TTC (s) & 2.0 & 5.0 & 1.0 \\
Center Lane TTC (s) & 6.0 & 8.0 & 1.0 \\
Right Lane TTC (s) & 3.0 & 6.0 & 4.0 \\
\midrule
Qwen3-14B Action & IDLE & IDLE & SLOWER \\
Gemma3-12B Action & IDLE & FASTER & SLOWER \\
\bottomrule
\end{tabular}
\caption{Columns correspond to the actions a human would most likely judge optimal for each scenario given the same prompt as LLM. Rows show the observation vector collected by the agent and, for the same observation, the action performed by each LLM. Qwen3-14B and Gemma3-12B choose conservative actions (IDLE or SLOWER) 
even when there is clear room for acceleration or when a lane change would be safer.}
\vspace{-10pt}
    \label{fig:llm-action-table}
\end{table}

We prompted the LLM using the prompt outlined in Figure~\ref{fig:llm-only_prompt}. Table~\ref{tab:llm_combined_metrics} reports the performance of the LLM-only agents. On the surface, success rates appear competitive, even exceeding the RL baseline in some environments (e.g., Gemma3-12B  reaches $94\%$ on \texttt{highway-fast}). However, this safety comes at the cost of efficiency: both models achieve very low speed scores (as low as $0.05$ for Gemma3-12B  on \texttt{highway-fast}). Examining behavior, LLM-only agents often choose to slow down or remain idle, exploiting a limitation of the simulator where driving at the lowest feasible speed reduces interactions with surrounding vehicles. Thus, while LLM-only agents can avoid collisions in certain settings, their overly conservative strategies highlight that success rate alone is not sufficient to characterize safe or effective driving. In particular, the prompt explicitly instructed the agent to maintain a high speed whenever safe, yet both models systematically ignored this requirement, suggesting that small LLMs either lack the reasoning capacity to process TTC observations in context or default to overly cautious behavior. These results indicate that the use of LLMs as the sole control policy may be inappropriate for this task, which motivates the need for hybrid approaches.

\section{Hybrid Design via LLM Reward Shaping}
\label{Hybrid Design}

\subsection{Hybrid RL+LLM Design}

We augment reinforcement learning with LLM-based reward shaping while keeping the DQN hyperparameters identical to those of the RL-only baseline (Figure~\ref{fig:rl-only_hyperparams}). The central idea is to incorporate an auxiliary signal from the LLM, which evaluates driving decisions in a human-like manner, into the reward computation. To achieve this, we design a custom prompt that queries the LLM about the quality of an agent’s action. The prompt presents the ego vehicle’s speed and the time to collision (TTC) with the leading vehicle in the current lane, both before and after an action is taken. The model is then asked to assign a scalar score between zero and ten, where higher values indicate safer and more efficient driving. The full text of this prompt is shown below.

\begin{tcolorbox}[colback=blue!5, breakable, colframe=blue!40!black, title=LLM Reward Shaping Prompt]
You are evaluating the behavior of an autonomous vehicle in its current lane.

TTC = Time To Collision — higher is safer.

--- BEFORE ACTION ---
\begin{itemize}
    \item Ego speed: \{prev\_obs[0, 0]\} m/s
    \item Current lane TTC: \{prev\_obs[2, 0]\} s
\end{itemize}

--- ACTION TAKEN ---
\begin{itemize}
    \item \{action\_str\}
\end{itemize}

--- AFTER ACTION ---
\begin{itemize}
    \item Ego speed: \{new\_obs[0, 0]\} m/s
    \item Current lane TTC: \{new\_obs[2, 0]\} s
\end{itemize}

Score this action from 0 (very unsafe or inefficient) to 10 (excellent decision). Prioritize avoiding collisions, maintaining approximately 30 m/s if safe, and smooth, human-like driving. Respond with only the numeric score.
\end{tcolorbox}

The total reward is defined as the sum of the environment reward and a shaped component derived from the LLM score,
\[
r^{\text{total}}_t = R(o_t, u_t) + \phi(s_t),
\]
where $R(o_t,u_t)$ is the environment reward, $s_t$ is the normalized LLM score, and $\phi$ is a shaping scheme that balances the two contributions. The environment reward follows the simulator’s definition,
\[
R(o_t, u_t) = a \cdot \frac{v_t - v_{\min}}{v_{\max} - v_{\min}} - b \cdot \mathbf{1}\{\text{collision at } t\},
\]
which encourages higher speeds while penalizing collisions. The simulator normalizes this reward into the $[0,1]$ range.

The input to the LLM is restricted to lane-focused features. Each observation $o_t \in \mathbb{R}^4$ consists of the ego speed $v_t$ and the minimum TTC values in the left, center, and right lanes, denoted by $\tau^L_t, \tau^C_t$, and $\tau^R_t$ respectively,
\[
o_t = \big[v_t, \tau^L_t, \tau^C_t, \tau^R_t\big].
\]
For prompting, we use only the ego speed and center-lane TTC before and after the action,
\[
x_t = (v_t, \tau^C_t), \qquad x_{t+1} = (v_{t+1}, \tau^C_{t+1}).
\]
Given the tuple $(x_t, u_t, x_{t+1})$, the LLM generates a quality score
\[
s_t = f_{\text{LLM}}(x_t, u_t, x_{t+1}) \in [0,10],
\]
which we normalize into the $[0,1]$ interval by setting $s_t \leftarrow 0.1 \, s_t$.

We explore three strategies for shaping. In the dense additive scheme, the normalized score is directly added to the environment reward,
\[
r^{\text{total}}_t = R(o_t, u_t) + \lambda \, s_t,
\]
with $\lambda$ controlling the influence of the LLM. Since both $R(o_t, u_t)$ and $s_t$ lie in $[0,1]$, the total reward is bounded by $[0,\,1+\lambda]$. We set $\lambda=1$ in our experiments. In the averaged fusion scheme, the two signals are combined with equal weight,
\[
r^{\text{total}}_t = 0.5 \, R(o_t, u_t) + 0.5 \, s_t,
\]
which keeps the reward in $[0,1]$ without requiring rescaling. Finally, in the centered shaping scheme, the LLM score is centered around its neutral midpoint $\mu=0.5$,
\[
\tilde{s}_t = s_t - 0.5, \qquad r^{\text{total}}_t = R(o_t, u_t) + \tilde{s}_t,
\]
so that $\tilde{s}_t$ lies in $[-0.5, 0.5]$. An affine transformation then rescales the resulting total reward back into $[0,1]$. This last method emphasizes deviations of the LLM’s evaluation from neutrality rather than its absolute magnitude.

\subsection{ Results for Hybrid RL+LLM Agent}

\begin{figure*}[t]
    \centering
    \includegraphics[width=\textwidth]{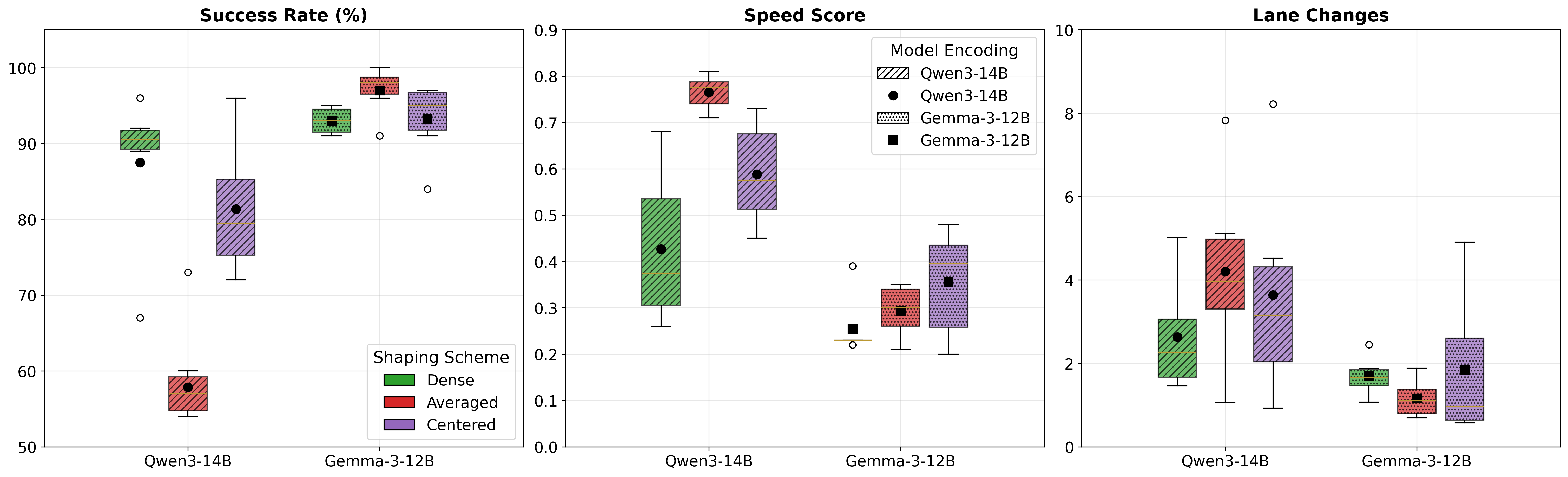}
    \caption{Comparison of reward shaping schemes across success rate (SR), lane changes (LC), and speed score. The figure highlights the spread and variability of outcomes under each scheme, complementing the aggregate results in Table~\ref{tab:hybrid_all_rewards}.}
    \label{fig:scheme_comparison}
    \vspace{-10pt}
\end{figure*}

We examine two LLMs, Gemma3-12B and Qwen3-14B, in their abilities to shape RL rewards in the three shaping schemes--dense, averaged, and centered. For each, we document success rate, speed score, and mean number of lane changes, as seen in Fig.~\ref{fig:scheme_comparison}). 
\paragraph{Dense shaping}
Under dense shaping, LLM scores are added directly to the environmental reward, providing an additional incentive for actions that LLM considers safe and reasonable. Because the scores are non-negative, this scheme never penalizes RL choices; it only amplifies positive feedback when the LLM approves a transition. Results of our hybrid approach using dense shaping are shown in (Table~\ref{tab:hybrid_all_rewards}). In aggregate, success rates  improve relative to the RL-only baseline in several settings: for example, on the \texttt{ highway} with 50,000 training steps, Gemma3-12B and Qwen3-14B achieve success rates of 95\% and 91\%, respectively, versus the RL-only baseline of 89\% that was shown in Table~\ref{tab:rl_combined_metrics}. This improvement demonstrates that the additional semantic signal effectively guides the agent toward safer behaviors.

However, this safety gain comes at a clear efficiency cost that reflects the incentive-only nature of the shaping signal. Speed scores sit between the two baselines: they are clearly higher than LLM-only but generally below RL-only levels. For instance, on \texttt{highway-fast} with 50,000 steps, Qwen3-14B's speed score falls to 0.26 compared to the RL baseline's 0.64. Lane-change counts also land between the extremes, remaining above the near-zero maneuvering of LLM-only yet typically fewer than RL-only. Gemma3-12B shows particularly conservative behavior, with as few as 1.07 changes on \texttt{highway-fast} with 20,000 steps. Overall, dense shaping boosts collision avoidance but biases the policy toward conservative throttle and fewer interventions, yielding a safety-efficiency trade-off (see corresponding subplots of Fig.~\ref{fig:scheme_comparison}).



\begin{table*}[t]
\centering
\resizebox{\linewidth}{!}{
\begin{tabular}{ccc*{9}{c}}
\toprule
\multicolumn{3}{c}{} &
\multicolumn{3}{c}{Dense} &
\multicolumn{3}{c}{Averaged} &
\multicolumn{3}{c}{Centered} \\
\cmidrule(lr){4-6}\cmidrule(lr){7-9}\cmidrule(lr){10-12}
Environment & Steps & LLM &
SR(\%) & Avg LC & Speed &
SR(\%) & Avg LC & Speed &
SR(\%) & Avg LC & Speed \\
\midrule
Highway       & 20k & Qwen3-14B   & 92  & 5.01 & 0.57 & 73  & 4.57 & 0.77 & 96 & 2.61 & 0.50 \\
Highway       & 20k & Gemma3-12B  & 95  & 1.44 & 0.23 & 100 & 1.89 & 0.35 & 94 & 4.91 & 0.48 \\
Highway       & 50k & Qwen3-14B   & 91  & 2.07 & 0.32 & 57  & 7.83 & 0.78 & 80 & 8.22 & 0.70 \\
Highway       & 50k & Gemma3-12B  & 95  & 2.45 & 0.23 & 99  & 1.15 & 0.26 & 97 & 0.68 & 0.22 \\
Highway-fast  & 20k & Qwen3-14B   & 96  & 3.26 & 0.43 & 60  & 3.37 & 0.71 & 87 & 1.85 & 0.45 \\
Highway-fast  & 20k & Gemma3-12B  & 93  & 1.07 & 0.23 & 98  & 1.45 & 0.26 & 91 & 3.06 & 0.42 \\
Highway-fast  & 50k & Qwen3-14B   & 89  & 1.46 & 0.26 & 57  & 5.11 & 0.73 & 79 & 4.52 & 0.60 \\
Highway-fast  & 50k & Gemma3-12B  & 91  & 1.55 & 0.22 & 96  & 0.71 & 0.21 & 96 & 0.57 & 0.20 \\
Merge         & 20k & Qwen3-14B   & 67  & 1.53 & 0.68 & 46  & 1.06 & 0.81 & 72 & 0.93 & 0.55 \\
Merge         & 20k & Gemma3-12B  & 93  & 1.88 & 0.23 & 91  & 1.05 & 0.34 & 84 & 1.24 & 0.44 \\
Merge         & 50k & Qwen3-14B   & 90  & 2.47 & 0.30 & 54  & 3.28 & 0.79 & 74 & 3.70 & 0.73 \\
Merge         & 50k & Gemma3-12B  & 91  & 1.77 & 0.39 & 98  & 0.69 & 0.34 & 97 & 0.62 & 0.37 \\
\bottomrule
\end{tabular}
}
\caption{Performance of Hybrid agent across reward variants. 
For each group of simulation environment, training steps, and LLM model, 
we report Success Rate (SR), mean number of lane changes (Avg LC), and speed score under {Dense}, {Averaged}, and {Centered} rewards. 
Across all schemes, Gemma3-12B consistently exhibits safer but slower behavior, 
while Qwen3-14B favors efficiency at the cost of safety, 
revealing persistent model-specific trade-offs regardless of shaping strategy.}

\label{tab:hybrid_all_rewards}
\vspace{-10pt}
\end{table*}



\paragraph{Averaged shaping}
Under averaged shaping (Table~\ref{tab:hybrid_all_rewards}), environment and LLM rewards receive equal weight, creating a balanced integration where both signals contribute equally to the total reward. This scheme yields variable outcomes that differ between the two LLM models. Gemma3-12B maintains excellent success rates, achieving up to 100\% on \texttt{highway} with 20,000 steps and consistently performing above 90\% across most environments. In contrast, Qwen3-14B's performance becomes more unstable, particularly with extended training: on \texttt{highway-fast} with 50,000 steps, Qwen3-14B's success rate drops to 57\%, well below both the RL baseline's 79\% and its own performance under dense shaping (89\%).

The efficiency metrics reveal a complex trade-off. Unlike in dense shaping, where Qwen3-14B was far slower than the RL-only baseline, using averaged shaping it maintained scores that surpassed  baseline levels, achieving 0.78 versus RL-only's 0.68 on \texttt{highway} with 50,000 steps, indicating that the agent maintains higher average speeds within the 20-30 m/s range. However, this comes with increased lane-change activity (7.83 changes compared to the RL baseline's 4.47), indicating more aggressive maneuvering that may contribute to the reduced success rates. Gemma3-12B exhibits the opposite pattern, maintaining very conservative behavior with low speed scores (0.26 on \texttt{highway} with 50,000 steps) and minimal lane changes (1.15), similar to its performance under dense shaping (see corresponding subplots of Fig.~\ref{fig:scheme_comparison}).

This model-dependent divergence reflects how different LLM architectures encode different risk preferences. When these preferences are equally weighted with the environment reward, the resulting learning dynamics vary substantially.



\paragraph{Centered shaping}
Under centered shaping (Table~\ref{tab:hybrid_all_rewards}), LLM scores are mean-shifted before integration, with values centered around 0.5 and then rescaled to maintain the [0,1] reward range. This approach aims to provide bias-corrected guidance by allowing both positive and negative deviations from the neutral midpoint. The results produce intermediate outcomes that fall between the patterns observed in dense and averaged shaping schemes.

Success rates show mixed results compared to the RL-only baseline. For example, on \texttt{highway} with 50,000 steps, Gemma3-12B reaches 97\%, compared to the baseline's 89\%, while Qwen3-14B only achieves 80\% . On \texttt{merge} with 50,000 steps, Qwen3-14B achieves 74\% versus the baseline's 73\%, showing comparable performance. 


The efficiency metrics show patterns intermediate between other shaping schemes. Rather than being much slower or faster, Qwen3-14B maintains speed scores of 0.73 on \texttt{merge} with 50,000 steps compared to the RL-only baseline's 0.74, and achieves 0.70 on \texttt{highway} with 50,000 steps versus the baseline's 0.68. Lane-change behavior varies considerably: Qwen3-14B exhibits high activity (8.22 changes on \texttt{highway} with 50,000 steps) that exceeds both dense shaping and the RL baseline (50,000 steps), while Gemma3-12B continues to show conservative patterns with minimal lane changes (0.68 on \texttt{highway} with 50,000 steps) and low speed scores.

\begin{figure} [t]
    \centering
    \includegraphics[width=0.48\textwidth]{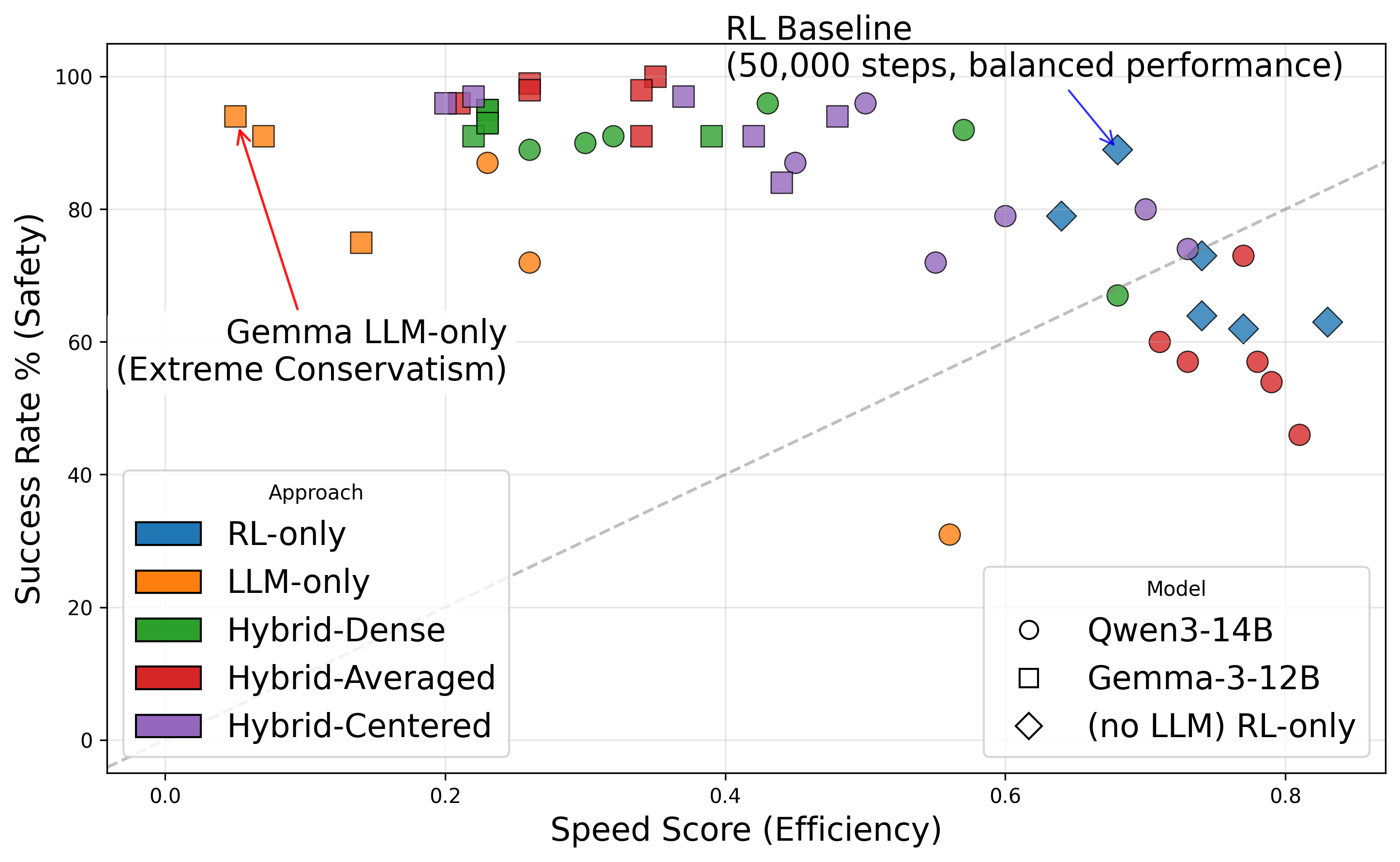}
    \caption{Safety–efficiency spectrum across approaches. Each point is one configuration (environment $\times$ training steps $\times$ model/scheme), plotted by success rate vs.\ speed score.}
    \label{fig:trade-off}
    \vspace{-10pt}
\end{figure}

The results suggest that centering may help balance the reward components, though the persistent differences between the results of Qwen3-14B and Gemma3-12B indicate that model-specific characteristics continue to influence behavior. Centered shaping appears to provide a middle ground between the other approaches, though the fundamental trade-offs between safety and efficiency metrics remain evident across all hybrid approaches(see ~Fig.~\ref{fig:trade-off}).

\section{Discussion}
\label{Discussion & Conclusion}

Motivated by the limitations of RL-only and LLM-only approaches to decentralized, partially observable highway driving, we examined utilization of hybrid approaches, where outputs from small LLMs are used to augment RL rewards. Our main takeaways are as follows: 


\begin{tcolorbox}[colback=blue!5, breakable,  colframe=blue!40!black, title=Main Takeaways]
\begin{itemize}
    \item Hybrid designs offer flexibility in balancing safety and efficiency in different aspects of highway driving.
    \item LLM-only approaches consistently show some level of conservative bias.
    \item Even under the same reward shaping method, LLMs perform differently.
    \item Merging is harder than highway driving.
\end{itemize}
\end{tcolorbox}


\subsubsection{Balancing Safety and Efficiency}

Our case study reveals a broad safety-efficiency trade-off spectrum across all examined approaches to autonomous highway navigation (Figure ~\ref{fig:trade-off}). RL-only agents achieve moderate success rates while maintaining reasonable speed scores and adaptive lane-changing behavior. LLM-only agents achieve high success rates in some environments but suffer from severely impaired efficiency. Hybrid RL+LLM approaches consistently fall between these extremes: dense shaping tends to fall closer to LLM-only conservatism, centered shaping produces intermediate outcomes, and averaged shaping shifts closer to either baseline depending on the LLM used. The variability of these hybrid results highlights the flexibility of reward shaping with local, small LLMs, enabling different priorities in highway driving.

\subsubsection{LLM-influenced Conservatism}
Despite explicit prompting to maintain efficient speeds, all LLM-influenced approaches (including our hybrid approaches) exhibit some level of conservative bias that prioritizes collision avoidance over driving efficiency. This pattern persists across different integration methods, suggesting such bias may originate from small LLMs' evaluation tendencies rather than implementation details.

\subsubsection{Model and Method Variability}
Our study also reveals substantial variability across LLMs. Gemma3-12B consistently achieves higher success rates but lower speed scores and fewer lane changes compared to Qwen3-14B across all evaluation schemes, as seen in Fig.~\ref{fig:scheme_comparison}. The largest discrepancy occurs under averaged shaping, where Qwen3-14B attains speed scores is on average 2.6x higher than Gemma3-12B's, but at the cost of success rates that are around 40\% lower. These results show that both choice of LLM and method of integration significantly influence driving behavior, even when prompts and training procedures are held constant.
 

\begin{figure}[t]
    \centering
    \includegraphics[width=0.48\textwidth]{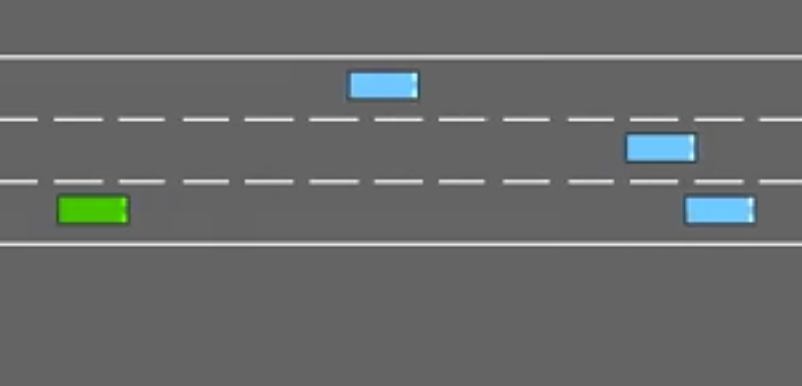}
    \caption{Other agents (blue) are notably ahead of the ego vehicle (green), creating fewer collision opportunities for a slow-moving ego vehicle.}
    \label{fig:traffic-artifacts}
\end{figure}
\subsubsection{Environmental and Simulator Limitations}

Performance also varies across test environments, with the \texttt{merge} scenario proving most challenging for all approaches. Success rates average 80\% on \texttt{merge} compared to 89\% for \texttt{highway}. This difficulty for \texttt{merge} likely stems from observation space limitations that omit crucial merging information that is needed for intent prediction in decentralized settings, such as turn signals or lane-specific positioning. Additionally, the simulation environment may introduce confounding factors that artificially inflate conservative strategies' apparent benefits. As seen in Figure~\ref{fig:traffic-artifacts}), agents maintaining slower speeds effectively reduce surrounding traffic density as faster vehicles overtake, creating fewer collision opportunities. Such affects complicate LLM-only result interpretation, where extremely conservative behaviors may exploit environmental characteristics rather than demonstrate effective driving strategies.


\subsubsection{Implications and Future Work}

Our case study highlights both the promise and the limitations of hybrid LLM-RL approaches to highway driving. Future research should investigate whether larger LLMs or alternative prompting strategies can mitigate the conservative bias observed in our results. Testing with richer observation spaces, including turn signals and relative positioning, would better support complex scenarios such as merging. Additionally, validation in more realistic simulation environments, free of traffic density artifacts could strengthen evidence for practical utility. Finally, exploring alternative reward shaping methods beyond the three schemes considered here may yield more effective ways to leverage semantic reasoning while maintaining highway driving efficiency.
\FloatBarrier

\bibliographystyle{IEEEtran} 
\bibliography{refs}

\end{document}